%% file: sensor2image_deepnet.tex
\documentclass[conference]{IEEEtran}
\usepackage{graphicx}
\DeclareGraphicsExtensions{.pdf,.jpg,.png,.svg}
\graphicspath{{figures/}}

\usepackage{footmisc}
\makeatletter
\let\old@ps@headings\ps@headings
\let\old@ps@IEEEtitlepagestyle\ps@IEEEtitlepagestyle
\def\confheader#1{%
	\def\ps@headings{%
		\old@ps@headings%
		\def\@oddhead{\strut\hfill#1\hfill\strut}%
		\def\@evenhead{\strut\hfill#1\hfill\strut}%
	}%
	\def\ps@IEEEtitlepagestyle{%
		\old@ps@IEEEtitlepagestyle%
		\def\@oddhead{\strut\hfill#1\hfill\strut}%
		\def\@evenhead{\strut\hfill#1\hfill\strut}%
	}%
	\ps@headings%
}
\makeatother

\confheader{%
	Published in: 2017 International Joint Conference on Neural Networks (IJCNN),
	14-19 May 2017, Anchorage, AK, USA
}

\ifCLASSINFOpdf
\else
\fi
\begin{document}
%
\title{Transforming Sensor Data to the Image Domain for Deep Learning - an Application to Footstep Detection}




\author{\IEEEauthorblockN{Monit Shah Singh\textsuperscript{\footnotemark{1}}\IEEEauthorrefmark{1},
		Vinaychandran Pondenkandath\textsuperscript{\footnotemark{1}}\IEEEauthorrefmark{2}\IEEEauthorrefmark{4},
		Bo Zhou\IEEEauthorrefmark{3}, \\
		Paul Lukowicz\IEEEauthorrefmark{1}\IEEEauthorrefmark{3} and 
		Marcus Liwicki\IEEEauthorrefmark{2}\IEEEauthorrefmark{4}}
	\IEEEauthorblockA{\IEEEauthorrefmark{1}TU Kaiserslautern, Germany}
	\IEEEauthorblockA{\IEEEauthorrefmark{2}MindGarage, TU Kaiserslautern, Germany}
	\IEEEauthorblockA{\IEEEauthorrefmark{3}DFKI, Kaiserslautern, Germany}
	\IEEEauthorblockA{\IEEEauthorrefmark{4}DIVA, University of Fribourg, Switzerland}
	\IEEEauthorblockA{$Monit\_Shah.Singh@dfki.de$, $vinaychandran.pondenkandath@unifr.ch$} \IEEEauthorblockA{$Bo.Zhou@dfki.de$, $Paul.Lukowicz@dfki.de$, $liwicki@cs.uni\mbox{-}kl.de$}
}


%


\IEEEoverridecommandlockouts
\IEEEpubid{\makebox[\columnwidth]{
		\copyright2017 IEEE \hfill} \hspace{\columnsep}\makebox[\columnwidth]{ }}
\IEEEpubid{\makebox[\columnwidth]{10.1109/IJCNN.2017.7966182~
		\copyright2017 IEEE \hfill} \hspace{\columnsep}\makebox[\columnwidth]{ }}

\maketitle



\begin{abstract}
Convolutional Neural Networks (CNNs) have become the state-of-the-art in various computer vision tasks, but they are still premature for most sensor data, especially in pervasive and wearable computing. A major reason for this is the limited amount of annotated training data. In this paper, we propose the idea of leveraging the discriminative power of pre-trained deep CNNs on 2-dimensional sensor data by transforming the sensor modality to the visual domain. By three proposed strategies, 2D sensor output is converted into pressure distribution imageries. Then we utilize a pre-trained CNN for transfer learning on the converted imagery data. We evaluate our method on a gait dataset of floor
surface pressure mapping. We obtain a classification accuracy of 87.66\%, which outperforms the conventional machine learning methods by over 10\%. 
\end{abstract}


\footnotetext[1]{These two authors contributed equally to this work.}

%
\IEEEpeerreviewmaketitle

\begin{figure*}
	\centering
		\includegraphics[width=15cm]{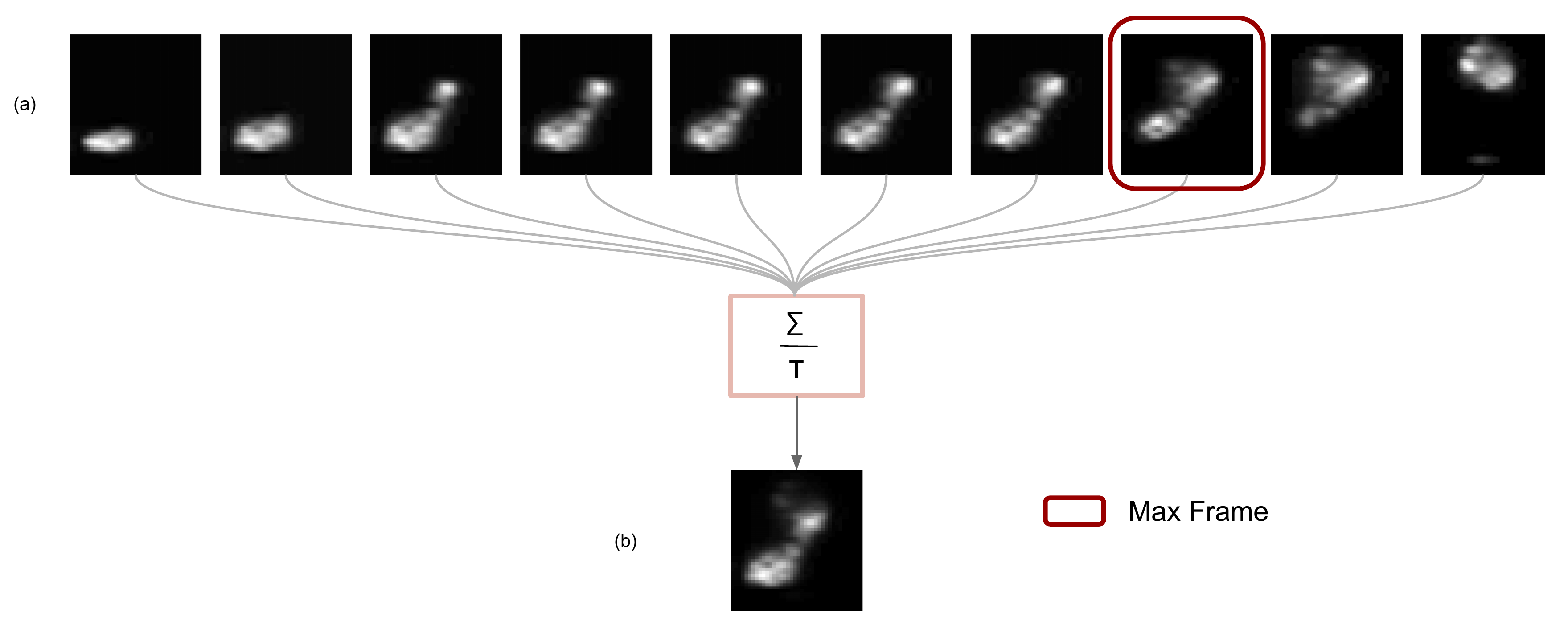}
	\caption{The step images obtained after modality transformation of pressure sensor data (a) all frames in a sequence of single step (walking direction is upwards), (b) average of all the frames in a sequence.}
	\label{fig:step_frames}
\end{figure*}

\input{Introduction}

\begin{figure}
	\centering
	\includegraphics[width=8cm]{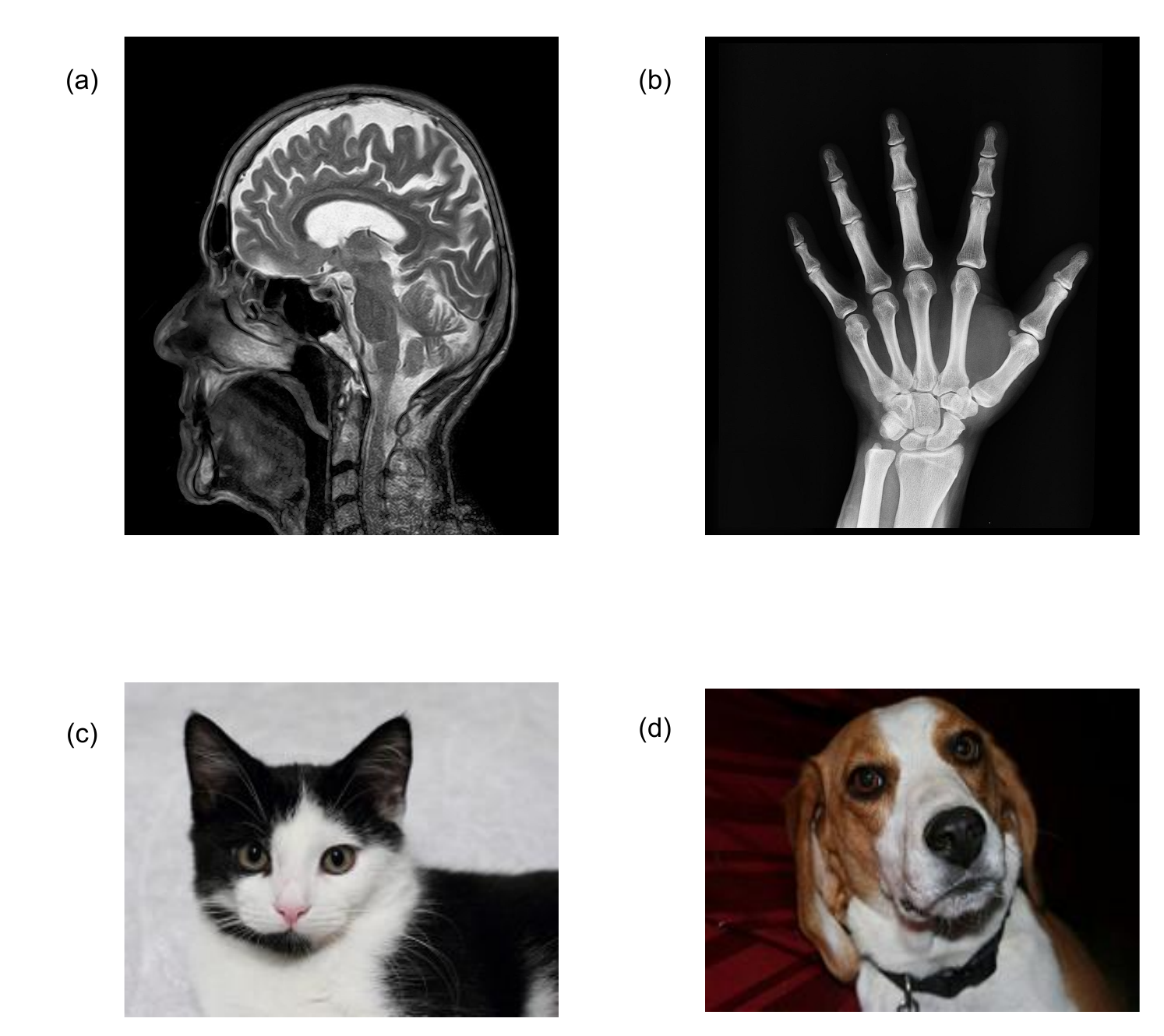}
	\caption{Examples of easily visually interpretable images, (a) Magnetic Resonance Imaging (MRI) scan, (b) X-ray scan, (c) Feline, (d) Canine }
	\label{fig:imagenet}
\end{figure}
\input{Pressure_sensor_data}
\input{Modality_transformation}

\begin{figure}
	\centering
		\includegraphics[width=8cm]{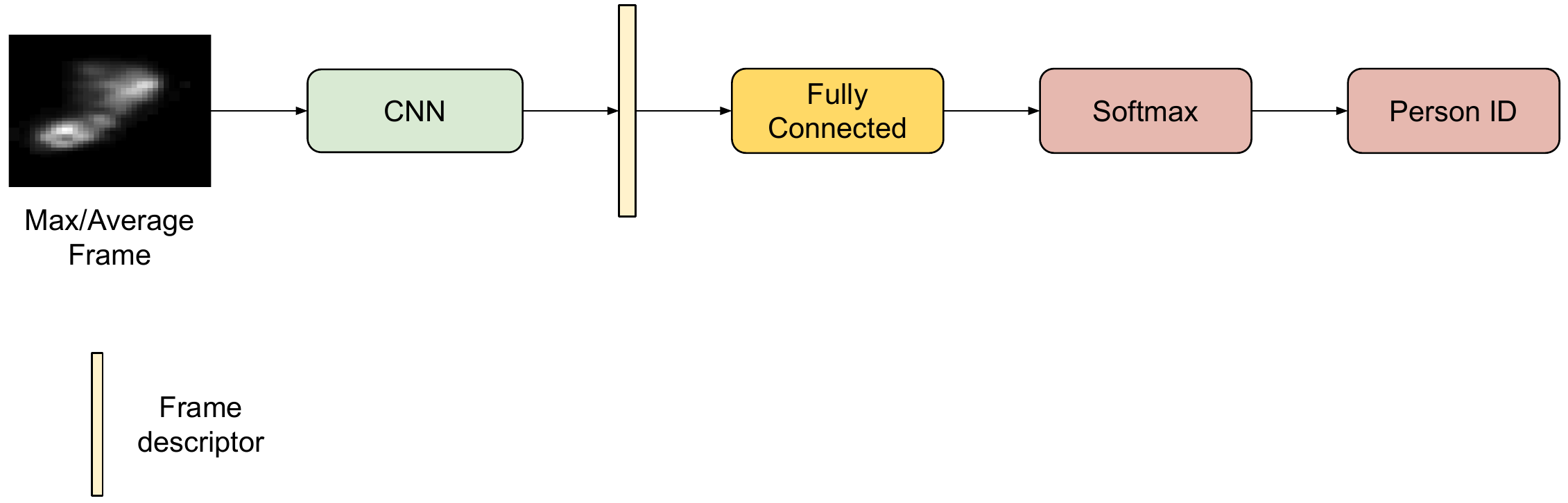}
	\caption{Schematic diagram of max and average(of sequence) frame classification pipeline.}
	\label{fig:ann_pipeline}
\end{figure}

\input{Inception-v3}

\input{Evaluation}

\input{Discussion}
\input{Conclusion}


\section*{Acknowledgment}

This research was partially supported by the Rheinland-Pfalz Foundation for Innovation (RLP), grant HiMigiac, HisDoc III project funded by the Swiss National Science Foundation with the grant number 205120-169618 and the iMuSciCA project funded by the EU (GA 731861). The authors would also like to thank all the experiment participants.
\par
The authors would like to thank Muhammad Zeshan Afzal and Akansha Bhardwaj for their valuable comments and the German Research Center for Artificial Intelligence (DFKI) and Insiders Technologies GmbH for providing the computational resources.



%
\bibliographystyle{IEEEtran}
\bibliography{deepsensor_ref}

\end{document}

%% file: Introduction.tex
\section{Introduction}
The presence of sensors in the ubiquitous environment has led to the production of an enormous amount of data. These sensors belong to diverse categories including planar pressure, thermal, optical, acoustic, and proximity modalities. 
They provide information for activity recognition and context-aware models that could be used for a wide range of applications such as automatic monitoring in the smart environment and wellness scene, computer-human-interaction, user experience, etc. 
There has been extensive research on wearable and pervasive sensors, which record and sense data in the context of human activities; ranging from physical activities (running, sleeping, walking, etc.), such as  monitoring gym exercises \cite{zhou2016never}, analyzing gait patterns \cite{tao2012gait}, to biological activities (breathing, eating, etc.), such as breathing detection \cite{corbishley2008breathing}, eating \cite{cheng2010active} and drinking arm gesture detection \cite{amft2005detection}. The task of extracting useful information from the raw sensor data has been performed using various machine learning and data mining techniques \cite{banaee2013data}. One of the many examples is the use of such methods in human activity recognition \cite{maurer2006activity}. Usually, when the raw sensor data is concerned, the feature extraction is done using numerical statistical features. These features have proven to be quite reliable in tasks related to classification, recognition and segmentation \cite{cheng2013smart}. 
\par

Deep learning has been recently proven to be extremely successful in various domains. Convolutional neural networks (CNNs) \cite{fukushima1979neural}, have already been applied to practical tasks by Le Cun et al. \cite{le1990handwritten}, they have recently risen in popularity after achieving superhuman accuracy on image classification tasks \cite{krizhevsky2012imagenet,he2015deep,zhang2016polynet,szegedy2016inception}. Recurrent neural networks (RNN) especially with Long Short-Term Memory cells (LSTM) \cite{hochreiter1997long} have been used to classify sequences \cite{graves2008unconstrained} and to recognize activities \cite{donahue2015long, du2015hierarchical} with varying degrees of success. Both, CNN and RNN have been used in combination to create systems which are capable of understanding images, and to provide temporal context to these individual images. 

\par

A limitation of these techniques, however, is the requirement of large amounts of labeled data to facilitate the training of these very deep networks. While the computer vision community has facilitated this requirement with large labeled datasets, such as, the ImageNet \cite{ILSVRC15} and MS-COCO \cite{lin2014microsoft} datasets for object recognition, classification, detection and captioning; for various other tasks not many labeled datasets exist because the scope can be very specific when compared to general images.

\subsection{Transfer Learning}
For many Computer Vision problems, the above-mentioned limitation can be bypassed by performing transfer learning, i.e., using labeled data from one domain and transferring the learned knowledge to a target domain. Transfer learning involves using the knowledge acquired on a specific task, and adapting this knowledge to a different, but related task. Caruana \cite{caruana1998multitask} first introduced the concept of multi-task learning, targeting the improvement in generalization by using the domain information of related tasks. A common scenario for transfer learning involves using a convolutional neural network trained on a very large dataset, and then further fine-tuning it on the target dataset which is relatively small in size. A pre-trained CNN is used for transfer learning by removing the last fully-connected layer and using the activations of the last hidden layer as the feature descriptors of the input dataset. The resulting feature descriptors are then used to train a classification model. Recently, transfer learning has been done on semantic segmentation of images  \cite{long2015fully}. The learned representations of fully convolutional networks like AlexNet \cite{krizhevsky2012imagenet}, VGGnet \cite{vgg} are transferred by fine-tuning the semantic segmentation task. Similarly, Li et al. explored the concept of transfer learning on images with limited semantic meanings which do not perform well for high level visual tasks. The use of large number of pre-trained generic object detectors improved performances on recognition tasks with simple classifiers like linear SVM \cite{li2010object}. The key advantage of transfer learning is that it removes the need to create a large dataset required to train the CNNs. Also, the time and computational resources needed to perform training on such a large scale is considerably high and thus, transfer learning benefits us by saving this additional cost.

\par
Conventionally, transfer learning has been performed on domains that are easily visually interpretable. We define a domain as being easily visually interpretable as follows:
\\\\
\emph{A domain is said to be easily visually interpretable if by looking at its visual representation, a human can extract relevant information and a sense of the meaning it conveys.}
\\\\
Additionally, if the data is conventionally visually interpreted, it belongs to the category of easily visually interpretable data. For example, images of everyday objects such as automobiles, animals, landscapes; documents, X-ray, and MRI scan data all belong to the category of easily visually interpretable images. In the context of these examples, an average human can easily learn to distinguish between different classes, e.g., several sub-species of animals (Fig. \ref{fig:imagenet}(c)(d)) or several models of automobiles \cite{russakovsky2015imagenet}. Also the classification of documents into categories such as legal, scientific, or historical is a trivial task for most people. Doctors or radiologists analyze and interpret MRI or X-ray data (See Fig. \ref{fig:imagenet}(a)(b)) to detect irregularities in healthy organs. The application of transfer learning on such images is feasible for mainly two reasons. Firstly, it is possible due to their nature of being visually meaningful and secondly, there are large datasets from the same domain available within the community to carry out the transfer learning tasks. 

\begin{figure}
	\centering
	\includegraphics[width=8cm]{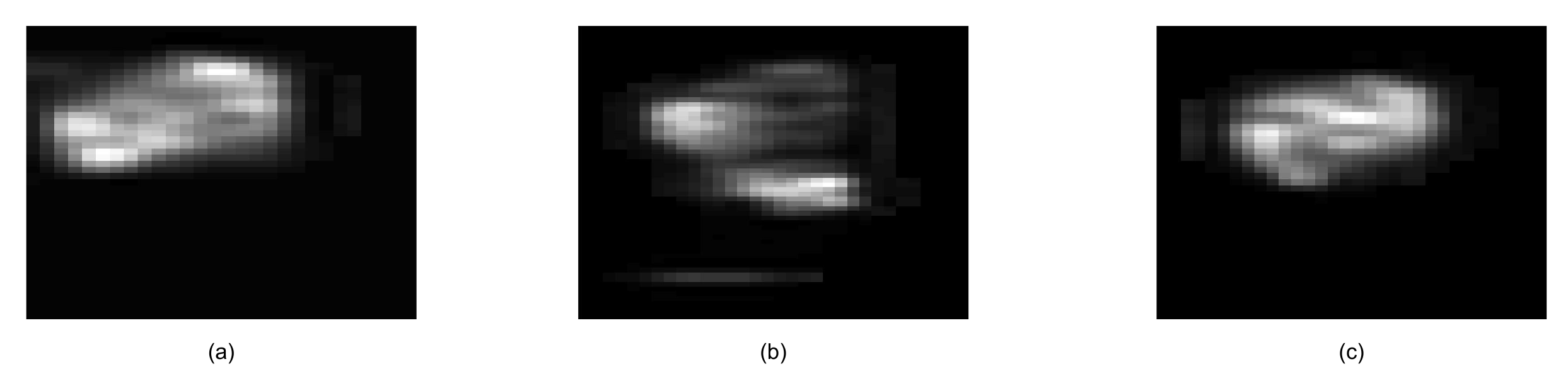}
	\caption{The transformed pressure sensor data corresponding to different moment of a step at different time. The heat maps (a) and (b) belong to the same person, (c) belongs to a different person.}
	\label{fig:similar_steps}
\end{figure}

However, there exist types of data, such as sensor data, which are not easily visually interpretable, and it is unclear if it would be possible to visually interpret them. Not visually interpretable data can be, for example, position updates of moving objects in location-based services, fluctuations in the stock market, medical experimental observations, or streaming sensor data.  An example is illustrated in Fig. \ref{fig:similar_steps} which shows the pressure mappings for a particular moment of a foot step. We can see that Fig. \ref{fig:similar_steps} (a) and (c) look similar but they belong to different persons. Similarly, Fig. \ref{fig:similar_steps} (a) and (b) look different but they belong to the same person. Hence, such sensor data is clearly not easily visually interpretable.

\subsection{Paper Contribution}
  In this paper, we introduce the idea of using the concept of transfer learning on a domain which is not easily visually interpretable. We carry out a shift procedure, which involves the shift from a domain which is not suitable for transfer learning to a domain on which the CNNs have been trained. This shift facilitates the use of transfer learning even for data which normally is not an ideal candidate for transfer learning. In order to show that such transformation is useful,  we extend the transfer learning methods on pressure sensor data. The raw sensor data is first transformed into a set of visual images and then used as an input dataset for a pre-trained convolutional network model. Thus, the core contributions of this paper are as follows:
\begin{itemize}
\item \textit{Modality transformation:} We transform non-interpretable data to the image domain and explore the effectiveness of deep neural networks. We observe that models which are pre-trained on the aesthetic and visually interpretable datasets like ImageNet, are powerful and accurate enough in terms of feature calculation that the artificially generated images are also recognized with high accuracy rates.

\item \textit{Unified feature extraction process:} Typically, the feature extraction process using conventional methods is customized for each unique application. In the case of sensor data, even with the same kind of sensors used for collecting the information, the feature extraction and data mining techniques vary depending on the target application of the data. The same pressure sensor has been used to cater to different applications \cite{cheng2016smart, SmartMat} but uses different feature extraction techniques. We provide a unified feature extraction process, which can be applied to the sensor data after conversion into the visual domain independent from its application.

\item \textit{Evaluation on pressure sensor data:} We evaluate our approach of modality transformation with pressure values of single steps as each person walks on a Smart-Mat \cite{SmartMat}, a fabric based real-time pressure force mapping system. The domain shift is carried out by transforming the pixel data values corresponding to the pressure exerted on the floor while walking (See Section \ref{pressuredata}), to the respective images. This information consisting of images serve as the input to pre-trained CNNs. With the application of our approach of modality transformation, we achieve a person identification accuracy of 87.66\% which significantly outperforms the state of the art (76.9 \%) (See Section \ref{sec:eval})

\end{itemize}

%% file: Pressure_sensor_data.tex
\section{Pressure Sensor Data}\label{pressuredata}
The dataset taken up for this research is taken from our previous work \cite{blind}, which consists of the step samples of 13 people who walk on a pressure sensitive matrix. Each person in each walking sequence records 2-3 steps. A minimum of 12 such samples is recorded for each person. The demography of participants varies in terms of height, weight, and shoe-size  from 155-195 (in cm), 64-100 (in Kg), and 37-45 (European size) respectively. This accounts for the high variance in the recorded data. Each walking sequence is an individual data sequence, labeled with a specific person ID which defines the class label for the CNN. Overall, 529 steps are recorded.

%% file: Modality_transformation.tex
\section{Modality Transformation}
	\label{sec:mod}
Modality transformation refers to the steps taken to convert the data from a source mode to a target mode. The purpose of such a transformation is to exploit the knowledge present in pre-trained models in the target mode to allow for easy classification. In this section, we describe the process of modality transformation on the example of pressure sensor data.

\begin{figure}
	\centering
		\includegraphics[height=6cm]{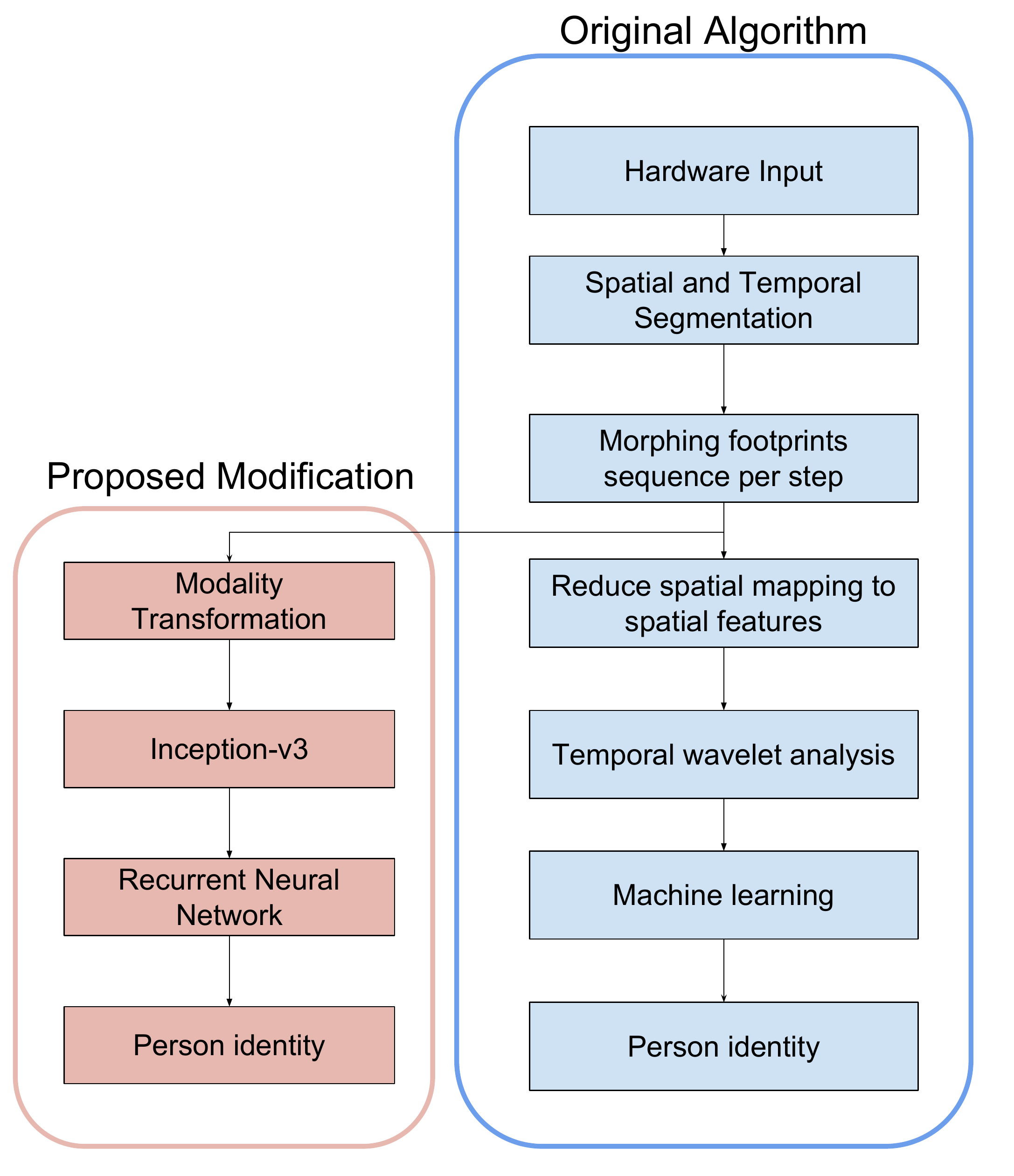}
	\caption{Original and proposed algorithm flow chart.}
	\label{fig:flowchart}
\end{figure}

%
In our specific case, the modality transformation of the raw data from sensors constitute the steps to convert the sensor mode into the visual mode in the form of images. The raw data from the sensor is a temporal sequence of $120 \times 54$ 2-dimensional pressure mappings. For the force-sensitive resistor fabric sensor, every sensing point's value is essentially the voltage potential measurement that is related to the pressure. We transfer these values linearly into a gray-scale color map, in which each pixel represents a sensing point, and brighter color corresponds to higher pressure. A complete step is a sequence of such pressure mapping frames; every frame corresponds to the moment of a step as shown in Fig. \ref{fig:step_frames}(a). The number of frames which comprise the entire step varies among people. Thus, it is important to segment each step along temporal dimension and find the individual moments for each step. It is noteworthy that, the idea of modality transformation is not limited to pressure data. For other matrix-sensors, a similar modality transformation strategy can be applied. The main idea of this paper is that after such a modality transformation, it is possible to apply transfer learning on the transformed data.

The modality transformation starts with the pre-processing of the raw data. Here, we first separate the steps from the background noise by converting each frame into a binary frame and applying an adaptive threshold. For the threshold, we sort the pixel values of the frame into a 10-bin histogram, and the threshold is decided as the center value of the next bin of the highest count bin.  However, this binarization process can be omitted for other pressure sensor data.

Next, we find the largest bounding box of all frames which encloses each individual step. It is, therefore, ensured that all the moments belonging to a same step will fit into that enclosing bounding box. Since, the bounding box is dynamically calculated for a step, the size of the enclosing box is different for different steps. However, within one step, all the moments are captured and extracted using the same sized bounding box. For the general modality transfer we suggest a position normalization in a similar manner, i.e., either cutting irrelevant parts with a bounding box or setting the center of the images to the mean position over all frames.

There can be many ways to convert the data from the source mode to a visual imagery mode. This depends on a number of factors; the dimensionality, range, heterogeneity or homogeneity, volume, noisiness. Ideally the best way is the one which transforms the source mode data into a form as close as possible to the target mode on which the CNN has been trained. In the following, we describe three ways in which we extract the images after fitting the bounding box. 

The transformation on the sensor data can be done with three strategies: max-frame, averaging, sequential analysis, giving us three possible images per data sequence (per step). 

For the first strategy, we capture the maximum frame out of the frame sequence of each sample, which corresponds to the frame with highest value of pixel sum, then convert it into the respective image and label it with the class ID. As shown in Fig. \ref{fig:step_frames} (the frame with red bounding box), in our dataset, we obtain one such image for every step, hence, in total we have 529 such images. 

For the second strategy, we average over all the frames in the sequence of a single sample and generate the corresponding image with averaged pixel values. This is visualized for footstep data in Fig. \ref{fig:step_frames}(b). This averaged frame carries the temporal information from all the moments of the step and should be more effective than the maximum frames for classification. 

For the third strategy, we use all the frames which form a temporal sequence of a step within each sample and transform them into the images. The sequence of frames capturing the moments of a single step are shown in Fig. \ref{fig:step_frames}. This carries the original raw values at each frame and provides more granularity than previous approaches for the feature set calculation.

%% file: Inception-v3.tex
\section{Architecture}

For transfer learning, our strategy follows the idea of transferring from the image classification task, i.e., using a pre-trained model from ImageNet or Coco-DB. Either the classification layer is removed or used as feature descriptor and a new classification layer is added. Thus the CNN is used as a fixed feature extractor.

The pre-trained CNN we use in our experiments is the Inception-v3 model from \cite{inceptionv3}. It is a CNN variant that focuses on improving computational efficiency along with performance. We choose this model for two reasons; with a Top-5 error of 3.58\%, it clearly performs extremely well on the ILSVRC-2012 classification benchmark and it requires relatively less computational resources to process the input. This ensures the possibility of applying such CNN models to real-time processing of high velocity sensor data.

\begin{figure*}
	\centering
	\includegraphics[width=18cm]{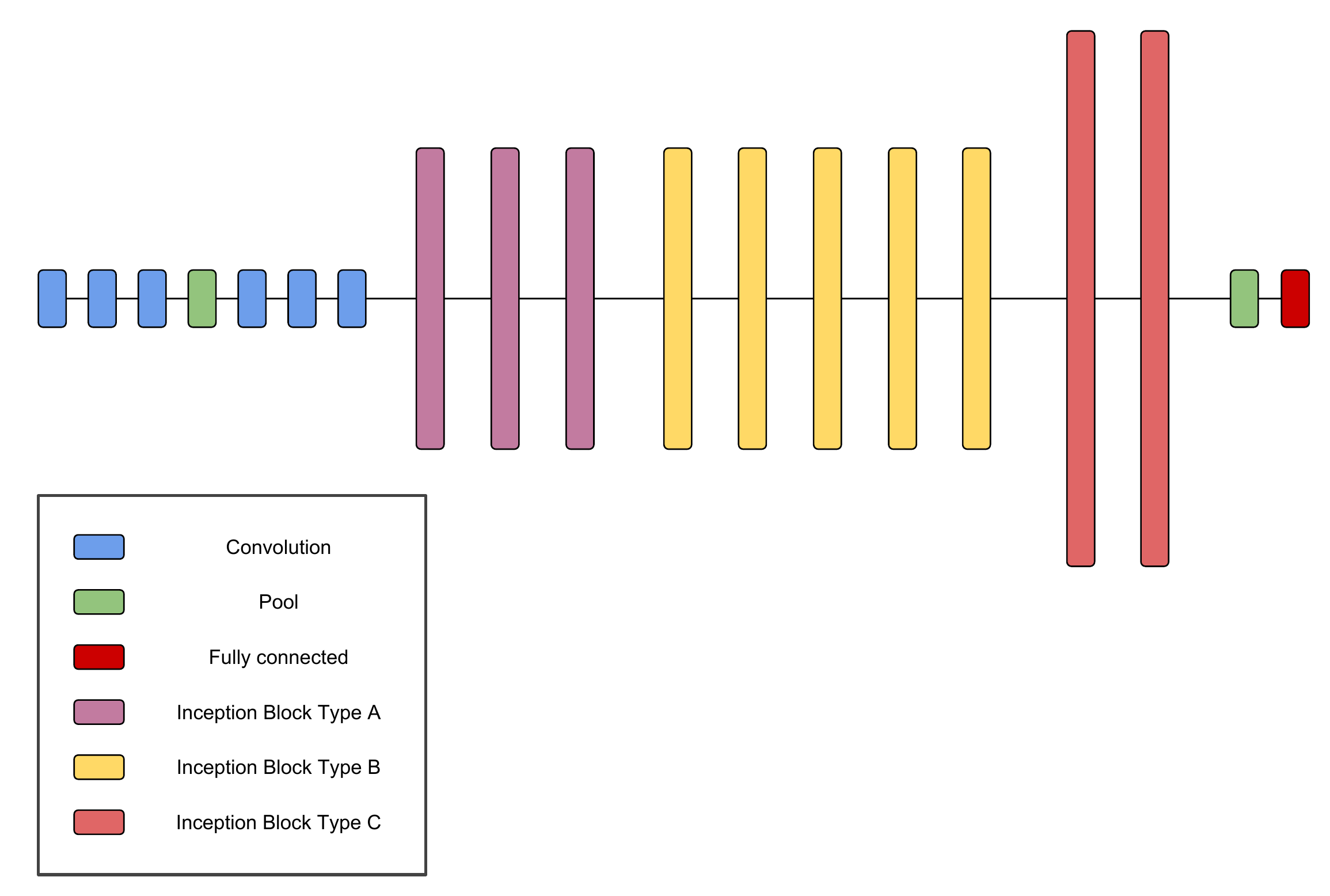}
	\caption{Simplified diagram of Inception-v3 cropped at the fully connected layer.}
	\label{fig:inception}
\end{figure*}

The Inception-v3 architecture consists of 3 convolutional layers followed by a pooling layer, 3 convolutional layers, 10 Inception blocks and a final fully connected layer. This results in 17 layers which can be learned by training the network on the data. We resize the images to the dimensions 229 x 229 as required by Inception-v3. We extract the activations from the fully-connected layer as shown in Fig. \ref{fig:inception}. This results in a 2048 dimensional output for each input. Each output can be interpreted as the descriptor for each frame in the sequence.

\begin{figure}
	\centering
		\includegraphics[width=9cm]{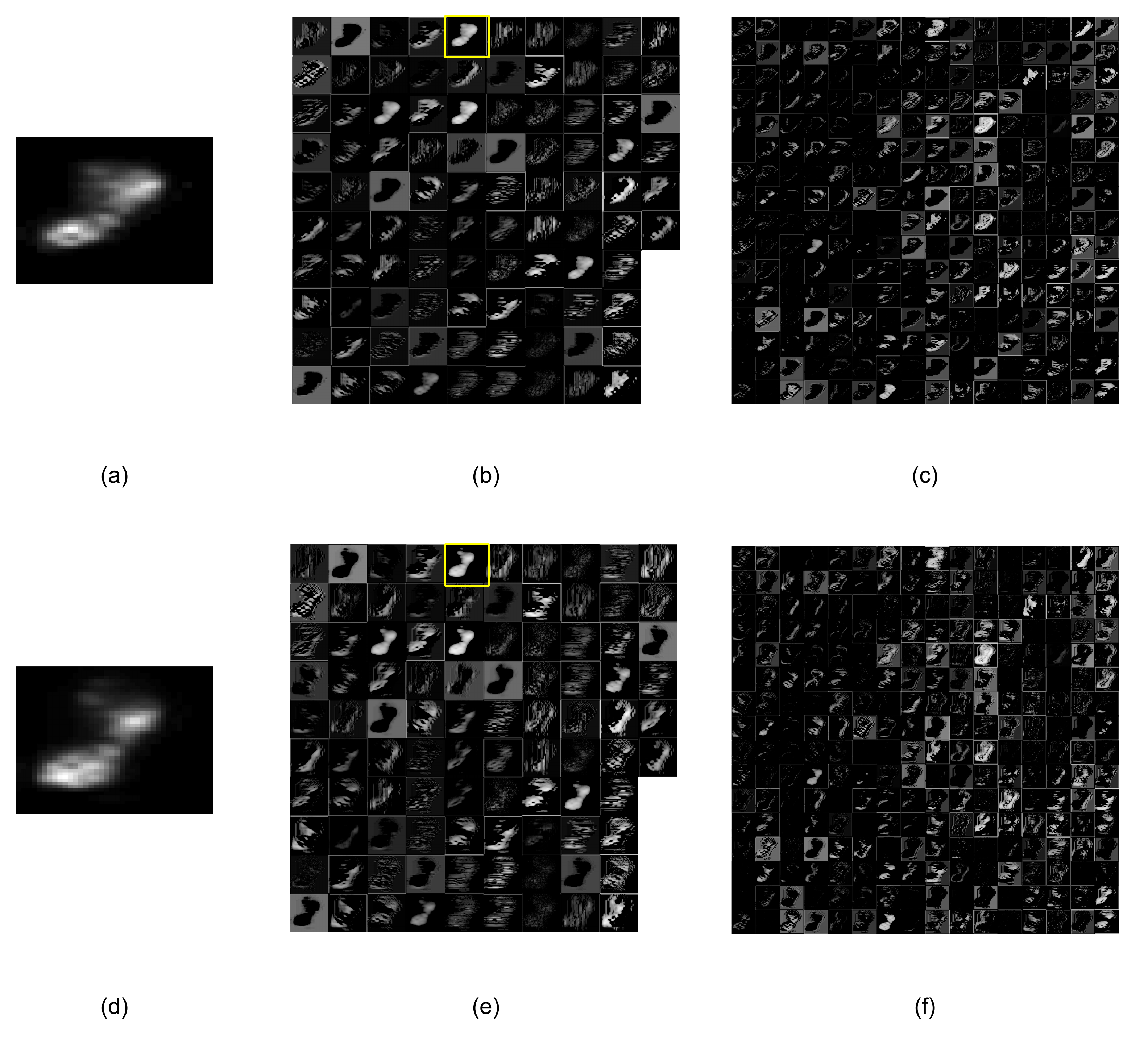}
	\caption{Visualization of information processing through different layers of AlexNet: (a) is the maximum frame and (d) the average frame of a single step sequence. (b) and (e) are the visualizations of the activation of the all filters in the first convolutional layer and (c) and (f) are visualizations from the second convolutional layer.    }
	\label{fig:max_conv_vis}
\end{figure}

The CNN will then be provided with the transformed input image (see Section \ref{sec:mod}) and resized to fit the CNN input size. The activations for the entire network
are computed by forward propagating the input through the network. As an example, Fig. \ref{fig:max_conv_vis} shows feature visualizations of all the filters present in the first and second convolutional layers of AlexNet \cite{deconv}. While it is possible to visualize the activations of deeper layers, typically they are harder to interpret.

\par

%% file: Evaluation.tex
\section{Evaluation}
	\label{sec:eval}
The evaluation procedure focuses on identifying a person from the footprints of individual steps. 
As shown in Fig. \ref{fig:step_frames}, the steps are present in sequences of the pressure mapping imageries of individual steps, which we use as the original dataset. The dataset includes overall 529 sequences from 13 participants. 

In our previous work \cite{blind}, a fast wavelet transform is applied to these sequences of steps to generate a single 336 dimension wavelet descriptor for each step. These features are then classified using a support vector machine classifier with a quadratic kernel \cite{cortes1995support}. This approach results in a classification accuracy of 76.9\%.
Our approach diverges after the steps are segmented and the flowchart for our proposed approach can be seen in Fig. \ref{fig:flowchart}. We generate the dataset of images which are maximum frame (sum of all pixels per frame), average frame (average of all the frames present in a sequence) and the set of all the frames forming a step; after that this set of images is passed through the pre-trained Inception model, for feature extraction, which upon classification gives the recognition results for person identification.

\subsubsection{Maximum Frames}
As described in Section \ref{sec:mod}, the maximum frame corresponds to the point at which the foot exerts maximum pressure on the ground. The pressure mat scans at the rate of 25 fps, thus a single frame corresponds to the information captured in 40 milliseconds. Our assumption is that the maximum frame of each step corresponds to the situation when a person's entire foot is on the ground, and hence, contains enough spatial information to be discriminative. 
\par
We evaluate our system by using these images for all the steps present in our dataset. These images are passed through the pre-trained Inception model and the activations of the fully-connected layer are used as image descriptors. To complete the classification task, the image descriptors are fed to a fully-connected layer, which computes the probability distribution over the different classes using \emph{softmax} activation function. The schematic diagram of the architecture followed for this approach is shown in Fig. \ref{fig:ann_pipeline}. This entire process is carried out with a 10-fold cross validation and repeated for 10 iterations. We calculate the average of the result obtained after each repetition. The final recognition rate after this method comes out to be 71.99\% as shown in Table \ref{tab:results}. 

\begin{table}[!t]
	\renewcommand{\arraystretch}{1.3}
	\caption{Person identification accuracies for different image sets and feature types.}
	\label{tab:results}
	\centering
	\begin{tabular}{|c|c|c|}
		\hline
		\bfseries Feature Type & \bfseries Image Set Type & \bfseries Accuracy \\
		\hline
		Wavelet Transformation & All sequences in a step & 76.9\%\\
		\hline
		Deep CNN & Maximum frame & 71.99\%\\
		\hline
		Deep CNN & Average frame & 78.41\%\\
		\hline
		Deep CNN + RNN & Complete step sequence & \bfseries87.66\%\\
		\hline
	\end{tabular}
\end{table}

\subsubsection{Average Frames}
The walking pattern of a person has a temporal component within it. This time dimension includes the way a person starts with engaging his/her foot on the floor, which begins with the heel strike and then carries on until the toe off. Within these stages, the way an individual exerts pressure on the floor varies from person to person. In order to accommodate this temporal information, we average over all the frames in the sequence of a single step and compute images corresponding to all the steps. The evaluation is carried out in the similar manner as with the maximum frames (as seen in Fig. \ref{fig:ann_pipeline}) with a $10 \times 10$ cross-validation and the average recognition rate is 78.41\% as seen in Table \ref{tab:results}.

\subsubsection{Image Sequences with Recurrent Neural Network}

\begin{figure}
	\centering
		\includegraphics[width=6cm]{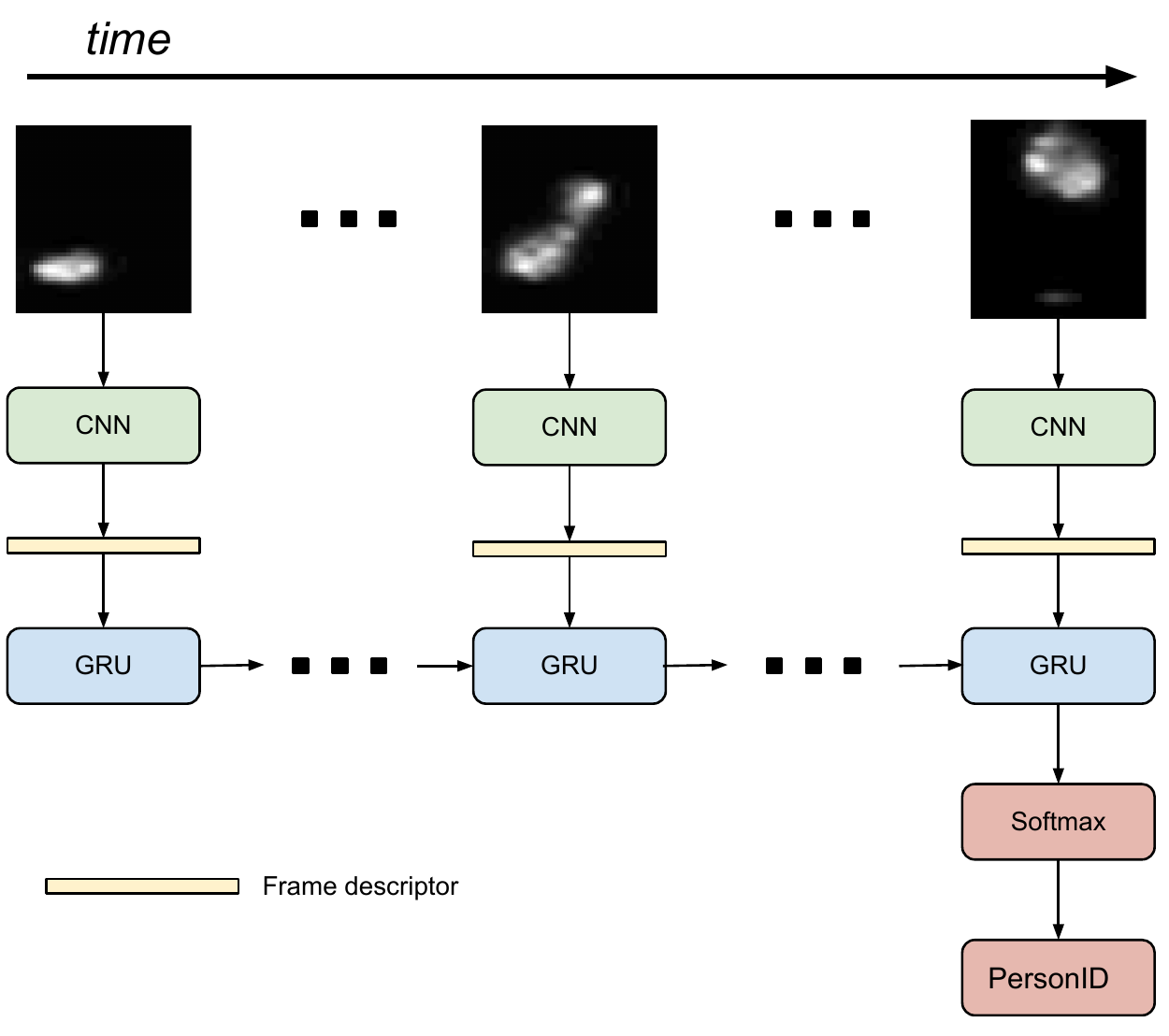}
	\caption{Schematic diagram of all frames in a single step classification pipeline.}
	\label{fig:rnn_pipeline}
\end{figure}

 In our experiments with the maximum and average frames, we observe that the average frames show an improvement over the  maximum frames. Even though the classification accuracy is improved by the information encoded in the average frames, a certain amount of information is lost by the averaging procedure. Hence, we experiment with using a RNN to classify the temporal sequence of each step. As shown in Fig.  \ref{fig:rnn_pipeline}, all of frames associated with a step are processed through the Inception-v3 model to extract a single descriptor for each frame. 
These descriptors are then fed one after another into a layer of Gated Recurrent Units (GRU) \cite{cho2014properties}, which generates a classification upon completing the sequence. We follow the same evaluation procedure outlined in the previous experiments and we obtain a classification accuracy of 87.66\% (Table \ref{tab:results}).

%% file: Discussion.tex
\section{Discussion}
We see that with the use of deep neural networks we achieve considerably better recognition results. The accuracy obtained with this method in case of average frames outperforms the baseline reported in our previous work which uses the conventional feature set for the same task \cite{blind}. It is noteworthy, because, the average frames are significantly more lossy than the data available to the conventional methods. When we use our method on all frames in the sequence of a step, we obtain an accuracy with an increase of over 10\% when compared to the accuracy of 76.9\% achieved by wavelet transformation. These results are directly comparable because both the methods evaluate a single step at a time for the person identification task.

Thus, we suggest three different approaches for generating the visual representations: max-frame, average-frame, and all frames in a sequence. While these approaches produce satisfactory results, finding a good representation for any other given sensor data can still be challenging if the data is of a very different nature. However, it is suggested to try these approaches even if the visual representations seem to be not convincing. A network transferred from an easily visually interpretable domain can still be able to distinguish the classes very well.

When visualizing the activations of the first and second convolutional layers (See Fig. \ref{fig:max_conv_vis}), we see a difference in the activations for the maximum and average frames in the first convolutional layers, despite both the input images being relatively similar. For example, it can be seen in \ref{fig:max_conv_vis} (b) and (e) [\emph{yellow}] that the shape of the foot is clearly distinguishable between the two images.

There can be many areas in which this approach can be implemented. Crowd-movement data generated from various sources can be used to model the traffic distribution over a geographical location; pollution particulate matter concentration over time can be visualized and considered as a time-series for prediction of air quality. Such distributed numerical data can be visualized on a geographical map or globe. The considered patterns can be assigned labels associated with some events of the world. For example, a pattern generated from the crowd-movement data in a city can signify the busiest parts of the city at any given time. In this case, a pre-trained CNN can be used to classify different types of city parts.



%% file: Conclusion.tex
\section{Conclusion and Future Work}

The use of transfer learning to extract features and solve typical AI tasks has been increasing over the past few years. The major consideration while doing transfer learning tasks is that the data in the source domain and the target domain are similar in terms of representation.

This paper explores the concept of transfer learning for the domains which are not directly eligible to apply transfer learning. We introduce the idea of transforming a non-visually interpretable problem domain to a visual domain, to leverage the effectiveness of pre-trained CNNs on visual data. To our knowledge, this kind of modality transformation of data with the intention of applying transfer learning techniques has not been carried out so far.
Additionally, we provide a unified feature extractor for sensor data which generally requires different feature extraction techniques for different applications. However, even though we apply our technique on the sensor data for a only single application, we believe that since the CNN for feature extraction is fixed and there is difference only in the way the input data is transformed, it should work for other sensor data applications as well.

This paper applies the introduced idea to a pilot dataset containing data from pressure sensors to perform a person identification task among 13 people. After evaluating the system with a pre-trained CNN as a feature extractor, we are able to achieve the average identification rate of 71.99\% and 78.41\% with maximum and average frame intensities respectively. We also explored the idea of analyzing the temporal information in the walking sequences, and applied the RNNs to exploit this additional dimension of time and hence achieve an average accuracy of 87.66\%.

This idea of modality transformation from one domain to another can be applied to other areas as well. This will specifically be beneficial in the cases of data for which there are no pre-trained models available. But, with this approach, if the data can be transformed into the target modality, transfer learning can be applied using the pre-trained models.


To further explore this concept, we would like to explore the effectiveness of this method on other types of sensors such as, accelerometers, gyroscopes, etc. Certain sensors do not present an obvious or intuitive way to be transformed into the visual mode. Therefore, it might be interesting to determine if it is possible to learn a transformation function from the source to the target mode.